\documentclass[letterpaper, 10 pt, conference]{ieeeconf}  

\IEEEoverridecommandlockouts                              
\title{\LARGE \bf
Flow6D: Discrete-to-Continuous Flow Matching for Efficient and Accurate Category-Level 6D Pose Estimation
}

\author{
\authorblockN{
Mingyu Mei$^{1,*}$, Li Zhang$^{2,*}$, Zibo Dai$^{1}$, Han Sun$^{3}$,
}
\authorblockN{
Xinyue Zhao$^{1}$, Huiliang Shen$^{1}$, Zaixing He$^{1,\dagger}$
}
\authorblockA{
$^{1}$Zhejiang University\quad
$^{2}$University of Science and Technology of China\quad
$^{3}$Shanghai Jiao Tong University
}
\authorblockA{
$^{*}$Equal contribution\quad
$^{\dagger}$Corresponding author
}
\authorblockA{
\href{mailto:mingyumei@zju.edu.cn}
{\texttt{mingyumei@zju.edu.cn}},
\quad
\href{mailto:zanly12138@gmail.com}
{\texttt{zanly12138@gmail.com}}
}
}

\usepackage{newfloat}
\usepackage{listings}
\usepackage{multirow}
\usepackage{amssymb}
\usepackage{bm}
\usepackage{booktabs}
\usepackage{enumerate}
\usepackage{url}
\usepackage[ruled,vlined]{algorithm2e}
\usepackage{amssymb,mathrsfs,amsmath}
\usepackage{makecell} 
\usepackage{multirow}
\usepackage{times}
\usepackage{epsfig}
\usepackage{graphicx}
\usepackage{amssymb}
\usepackage{booktabs}
\usepackage{url}
\usepackage{color}
\usepackage[english]{babel}
\usepackage{blindtext}
\usepackage{xspace}

\usepackage[backend=bibtex, style=numeric]{biblatex} 
\definecolor{lightblue}{RGB}{80, 160, 220}
\usepackage{hyperref}

\makeatletter

\newcommand{\Rmnum}[1]{\expandafter\@slowromancap\romannumeral #1@}
\makeatother

\begin{document}
\maketitle
\thispagestyle{empty}
\pagestyle{empty}
\begin{abstract}
6D pose estimation is a key task in computer vision and embodied AI, widely used in robotic manipulation, augmented reality, etc. Existing methods directly regress in a high-dimensional continuous space, facing two key challenges in category-level pose estimation: limited accuracy due to noise and local optima, and inefficient search over an infinite space that hinders real-time performance. This paper proposes Flow6D, a hierarchical flow matching framework with a two-stage \textit{discrete latent space localization-continuous pose regression} strategy. Rotation and translation parameters are first discretized into bins, with a discrete flow matching model locking the latent space around the true pose to reduce search complexity. Then, by sampling in the latent space, a continuous flow matching model predicts local pose residuals to optimize the estimate and regress to an accurate pose. The framework also naturally extends to articulated objects, outperforming state-of-the-art methods on synthetic and real datasets with real-time inference at 70 FPS. Project website: \color{lightblue}{https://flow6d.github.io/}.

\end{abstract}

\section{INTRODUCTION}

Object pose estimation~\cite{peng2022self, zhang2025r} is a core task in computer vision and embodied AI, with applications in robotics manipulation~\cite{kong2023design}, augmented reality~\cite{hidayah2019walking}, human-computer interaction~\cite{kastritsi2024passive}, and scene understanding~\cite{cordts2016cityscapes}. Accurate 6D pose (3D rotation and 3D translation) estimation enables precise robot–environment interaction, supporting stable grasping of rigid objects and flexible manipulation of articulated objects. As a fundamental perception capability, reliable pose estimation is essential for the practical deployment of intelligent systems in complex real-world environments.

Despite substantial progress in instance-level and category-level pose estimation methods, existing solutions still face notable limitations when addressing  pose estimation for rigid and articulated objects in complex scenarios. Most approaches adopt direct regression to predict pose parameters in continuous space, which is sensitive to observation noise and prone to local optima, leading to degraded accuracy under occlusion or partial observations. This issue is particularly pronounced for articulated objects, where pose errors of individual parts may accumulate and propagate through kinematic structures. Moreover, directly optimizing pose parameters in an unbounded continuous space requires searching over a vast solution domain, resulting in high computational cost and prolonged inference time. Consequently, existing methods often struggle to balance pose accuracy and real-time performance.

\begin{figure}[t!]
    \centering
    \includegraphics[width=1.0\linewidth]{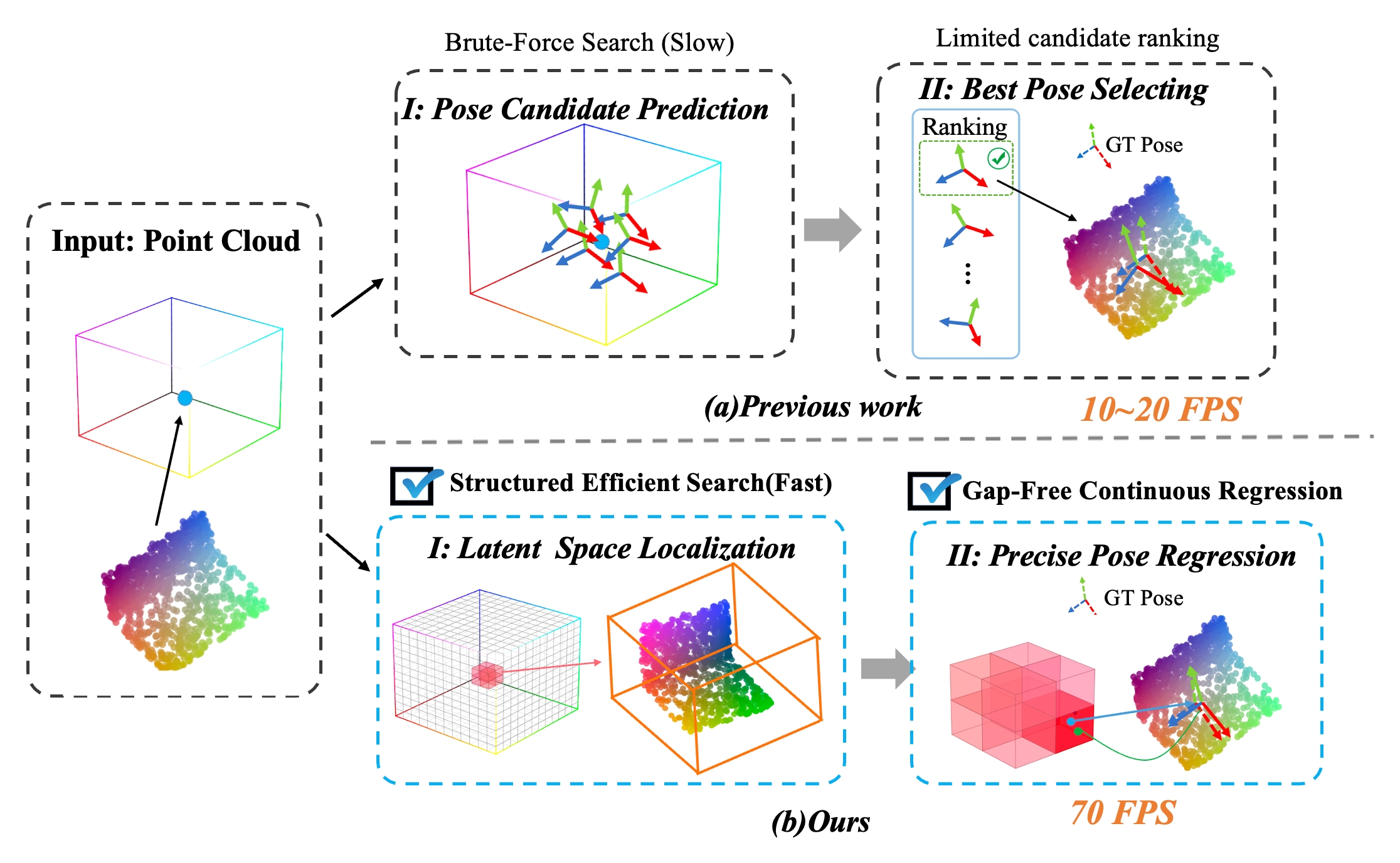}
    \caption{\textbf{Comparison between prior candidate-based pose estimation pipelines and our approach. (a) Previous methods depend on brute-force and limited candidate ranking, incurring high cost and accuracy limitations. (b) Our method adopts latent-space localization and continuous pose regression, achieving higher accuracy and faster speed.}}
    \label{fig:pipeline}
    \vspace{-1.3em}
\end{figure}

To address the aforementioned challenges, unlike prior work~\cite{zhang2023generative}, which relies on diffusion for pose candidate sampling, likelihood-based filtering, and ranking of remaining candidates, resulting in slow brute-force search and relying on the candidate prediction. As shown in Fig.~\ref{fig:pipeline}(b), our method is centered on a two-stage \textit\textbf{{Latent Space Localization-Precise Pose Regression}} pipeline, which achieves precise pose modeling from discrete to continuous representations through structured efficient search and gap-free continuous regression, enabling real-time performance at 70 FPS.

In the latent space localization stage, we utilize a discrete flow matching ~\cite{gat2024discrete} model for latent space prediction: rotation and translation parameters are discretized into a finite number of bins according to preset step sizes, and the discrete flow matching process is used to learn the probability distribution of pose parameters in the bin space, thereby quickly locking in the approximate latent space of the true pose. This classification-based discretization approach not only effectively mitigates the search complexity inherent in infinite continuous spaces but also harnesses the strong distribution modeling capacity of discrete models and the intrinsic stability of classification tasks. By doing so, it avoids the tendency of direct regression to get trapped in local optima within continuous space, thereby establishing a robust and reliable foundation for subsequent precise optimization.

In the subsequent pose regression stage, a continuous flow matching~\cite{lipman2022flow} model maps from the structured latent space to an accurate 6D pose. Using the initial estimate, a localized residual pose space is defined, within which the model predicts fine-grained rotation and translation residuals through continuous sampling and deterministic transport. Concentrating on high-probability regions allows efficient compensation for discretization artifacts and enables direct regression to precise coordinates, achieving seamless transition from latent representation to exact 6D pose.

Our contributions can be summarized as follows:
\begin{itemize}
    \item We propose \textbf{Flow6D}, \textcolor{black}{a hierarchical framework that decomposes 6D pose estimation into a two-stage discrete-to-continuous flow matching process.} Discrete flow matching first identifies high-probability bins in a 6D latent space, while continuous flow matching optimizes the pose within high-confidence subspaces.

    \item Our framework conducts pose estimation for rigid and articulated objects by combining structured bin discretization with seamless local residual optimization. \textcolor{black}{This hierarchical strategy effectively constrains the infinite search space and mitigates local optima across diverse object categories and scenes.}

    \item Extensive experiments on diverse synthetic and real-world datasets demonstrate the effectiveness, generalization, and real-time performance of our proposed method, which outperforms state-of-the-art baselines.
\end{itemize}

\section{Related Work}  \label{sec:Related_Work}

\subsection{Category-level 6D  Pose Estimation}

Category-level 6D pose estimation aims to generalize to intra-class unseen objects~\cite{zhang2025u,zhang2023generative}, evolving into two core paradigms: shape prior-based~\cite{wang2021category,zou20226d} and shape prior-free methods~\cite{lin2021dualposenet,tu2025language}. The former learns category-specific shape priors (e.g., NOCS space~\cite{wang2019normalized}) offline and solves poses via registration algorithms, achieving high accuracy but relying heavily on CAD models with limited generalization. The latter offers greater flexibility by directly mining geometric and semantic features without relying on priors. 
Recent advances~\cite{li2025gce,lin2023vi, chen2024secondpose} explore improved geometric representations, SE(3)-consistent feature learning, and scalable training paradigms, enhancing robustness to intra-class variations and narrowing the synthetic-to-real domain gap.
Additionally, hierarchical modeling for articulated objects~\cite{liu2023categoryRL,zhang2025r}, has expanded the paradigm’s applicability to complex objects.

\subsection{Diffusion Model-Based Methods}
Diffusion-based pose estimation~\cite{liu2025diff9d,huang2025raypose,xu20246d} has recently emerged as a promising direction for uncertainty-aware pose modeling, with existing methods mainly falling into continuous and discrete diffusion paradigms.
Continuous diffusion methods~\cite{jiang2023se,liu2025monodiff9d} iteratively denoise in continuous pose spaces, either operating on SE(3) to maintain geometric consistency~\cite{jiang2023se} or conditioning on LVM features for robust category-level 9D estimation~\cite{liu2025monodiff9d}. Despite their suitability for fine-grained refinement, they typically rely on dense sampling and many denoising steps, resulting in high inference cost. Acceleration techniques such as DDIM~\cite{song2021denoising} alleviate but do not eliminate this limitation.
Discrete diffusion approaches~\cite{wang2024di2pose,austin2021D3PM} discretize pose parameters into bins or codebooks to streamline search complexity and enable efficient coarse localization. However, they face a rigid efficiency–accuracy trade-off: finer discretization enhances precision but escalates computational overhead and regression difficulty. To break this bottleneck, we propose a discrete-to-continuous flow matching framework that achieves high-precision pose estimation with efficient inference, overcoming the limitations of prior diffusion-based methods.

\section{METHOD} \label{sec:method}

\begin{figure*}[t!]
    \vspace{-0.8em}
    \centering
    \includegraphics[width=0.85\linewidth]
    {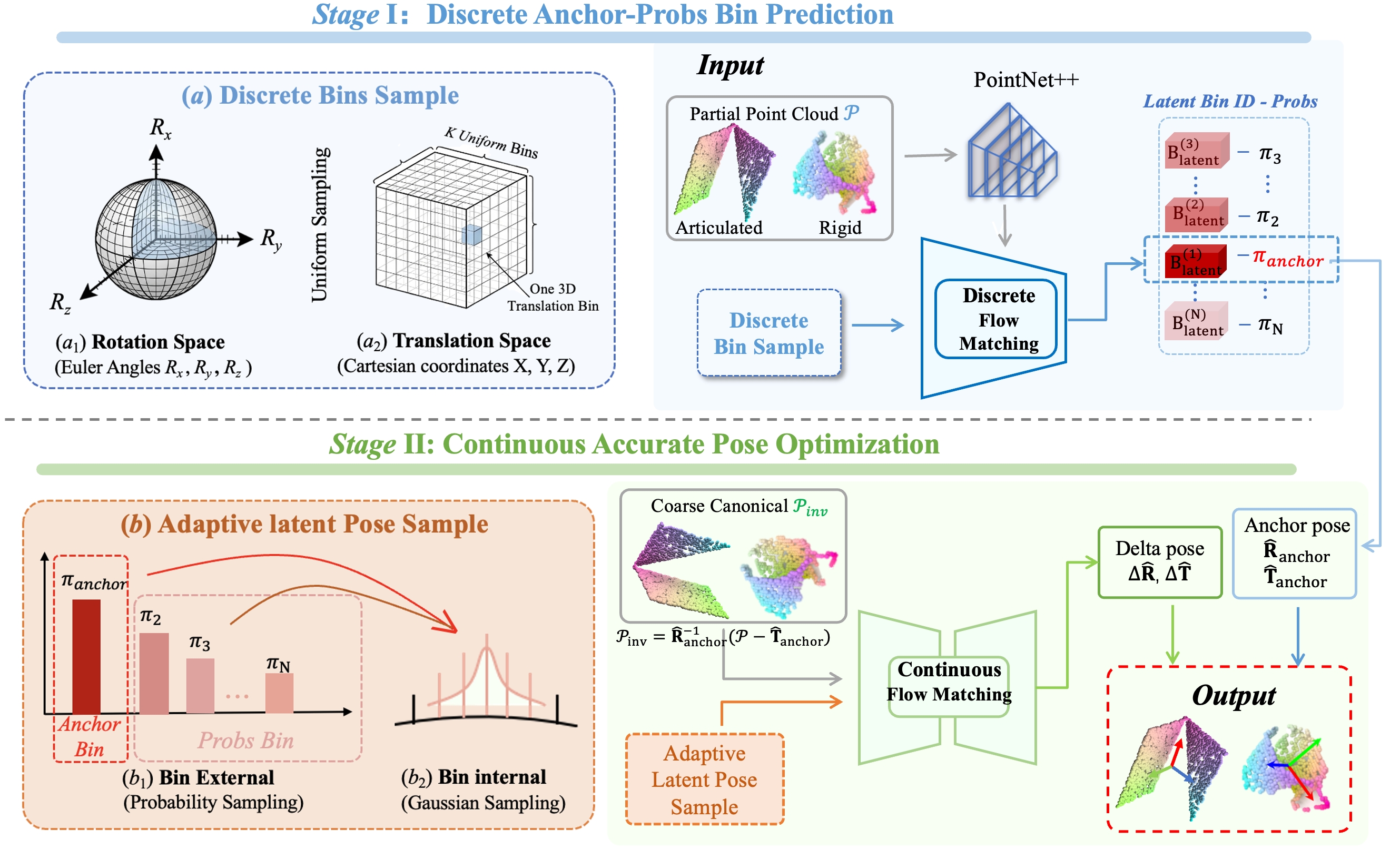}
    \vspace{-0.5em}
    \caption{\textbf{Overview of our two-stage pose estimation framework. Stage I performs discrete anchor-bin probability prediction by uniformly sampling rotation and translation spaces and selecting an anchor pose via discrete flow matching. Stage II optimizes the pose via continuous flow matching with adaptive latent pose sampling, enabling fine-grained, gap-free pose regression for accurate final estimation.
} }
    \vspace{-0.8em}
    \label{fig:framework}
\end{figure*}

As illustrated in Fig.~\ref{fig:framework}, Flow6D employs a hierarchical architecture that bridges discrete localization with continuous refinement. Initially, a PointNet++~\cite{qi2017pointnet++} encoder extracts features from the point cloud $\mathcal{P}$ to condition a discrete flow matching module. And DFM module independently models probability distributions across six pose dimensions, effectively constraining search space to high-confidence latent bins.

Subsequently, continuous flow matching regresses sub-bin residuals by modeling anchor-probs latent pose sampling, achieving precise pose recovery beyond discretization limits. For articulated objects, our method can naturally extends to articulated objects via a joint-centric strategy~\cite{zhang2025r}to predict the poses of the child parts.

\subsection{Discrete Bins Sampling}

As shown in Fig.~\ref{fig:framework}(a), to overcome the infinite dimensionality and high computational cost of continuous 6D pose estimation, we discretize the pose space into a finite set of latent bins~\cite{wang2019normalized}. This reformulates the original continuous search problem as an efficient structured latent search, providing a reliable initialization for subsequent refinement.

For translation, we sample the dominant spatial distribution from the category-level dataset and uniformly divide each axis ($x, y, z$) into $K$ bins. For rotation, We parameterize rotation using \textbf{Euler angles $(R_x, R_y, R_z)$}, uniformly discretized into $K$ bins. Compared to high-dimensional representations like \textbf{6D rotation representation or 9D rotation matrices}, Euler angles maintain only 3D structure, significantly reducing the denoising complexity for flow matching. Furthermore, unlike \textbf{quaternions or axis-angle} representations, which often lead to non-uniform or discontinuous bin partitions. Euler angles allow for intuitive, uniform discretization that aligns with point cloud observations. Potential gimbal lock issues are mitigated through boundary bin merging.

By discretizing each dimension into $K$ uniform bins, each of the 6D pose dimensions is independently mapped to a finite set $\mathcal{B}^i = \{B_1^i, \ldots, B_K^i\}$, where the superscript $i$ indexes the $i$-th dimension. This reformulates pose estimation as a structured bin classification problem, streamlining search complexity and bypassing exhaustive exploration.

\subsection{Latent Space Modeling via DFM}

\paragraph{Discrete Latent Bin Training}
We leverage discrete flow matching (DFM)~\cite{gat2024discrete} to model the conditional probability distribution over a discretized 6D pose space $\mathcal{B}$ given an input point cloud $\mathcal{P} \in \mathbb{R}^{N \times 3}$. This enables efficient identification of the bin best aligned with the true pose by learning a time-evolving discrete probability flow $p_t(b \mid \mathcal{P})$, where $t$ denotes the timestamp, $b$ denotes a categorical state within the discretized 6D pose space $\mathcal{B}$. The flow interpolates from a uniform source distribution $p_0(b \mid \mathcal{P})$ to a target distribution $q(b \mid \mathcal{P}) = \delta_{b^\ast}(b)$, with $b^\ast$ denoting the ground-truth bin. Compared to diffusion models with iterative sampling and slow convergence, DFM learns a deterministic velocity field for faster, more stable, and more robust pose estimation.

For any source–target bin pair $(b_0, b_1)$, the conditional path is defined as a convex combination of point masses:
\begin{equation}
p_t(b \mid \mathcal{P}, b_0, b_1) = (1 - \kappa_t)\, \delta_{b_0}(b) + \kappa_t\, \delta_{b_1}(b),
\label{eq:conditional_path}
\end{equation}
where $\kappa_t$ is a monotonically increasing schedule with $\kappa_0 = 0$ and $\kappa_1 = 1$. We adopt a quadratic schedule $\kappa_t = t^2$, which emphasizes exploration of the initial distribution at early times and sharpens focus on the target bin as $t \to 1$.

The full flow path is obtained by marginalizing over all bin pairs under a learned coupling distribution $\pi(b_0, b_1 \mid \mathcal{P})$:
\begin{equation}
p_t(b \mid \mathcal{P}) = \sum_{b_0 \in \mathcal{B}} \sum_{b_1 \in \mathcal{B}} p_t(b \mid \mathcal{P}, b_0, b_1) \, \pi(b_0, b_1 \mid \mathcal{P}).
\label{eq:flow_path_marginal}
\end{equation}
This formulation induces a discrete velocity field that drives deterministic evolution from random initialization to high-confidence bins, balancing search efficiency and accuracy.

To train the model, we use a hybrid loss combining generalized Kullback–Leibler (KL) divergence and mean squared error (MSE). The KL term aligns the predicted flow with the optimal transport path:
\begin{equation}
\begin{split}
\ell^i(b_1, b_t, t) = -\frac{\dot{\kappa}_t}{1 - \kappa_t} \big[ &
    p_{1|t}(b_t^i \mid b_t) - \delta_{b_1^i}(b_t^i) \\
    & + \bigl(1 - \delta_{b_1^i}(b_t^i)\bigr) \log p_{1|t}(b_1^i \mid b_t)
\big],
\end{split}
\end{equation}

averaged across the $d=6$ pose dimensions to yield $\mathcal{L}_{\text{KL}}$. The MSE term supervises the expected bin indices $\hat{b}^i = \sum_k p_k^{(i)} \cdot k$ against ground-truth indices $b^{i*}$:
\begin{equation}
\mathcal{L}_{\text{MSE}} = \frac{1}{d} \sum_{i=1}^d (\hat{b}^i - b^{i*})^2.
\end{equation}

The total training objective is a weighted sum:
\begin{equation}
\mathcal{L}_{\text{DFM}} = \alpha \mathcal{L}_{\text{KL}} + \beta \mathcal{L}_{\text{MSE}},
\end{equation}
In all our experiments, we set $\alpha = 1$ and $\beta = 0.1$, placing primary emphasis on the KL divergence term to ensure probabilistic calibration, while using the MSE loss as an auxiliary objective to stabilize regression accuracy.

\paragraph{Discrete Flow Inference}
During inference, we utilize the backward-time velocity field of   Discrete Flow Matching~\cite{gat2024discrete} to generate samples from noise (\(t=1\)) to data (\(t=0\)). The backward-time update follows a first-order Euler discretization of the discrete probability flow:
\begin{equation}
    b^i_{t-h} \sim \delta_{b^i_t}(\cdot) - h\, u^i_t(b^i_t, b \mid \mathcal{P}),
    \label{eq:backward_sampling}
\end{equation}

where the backward-time generating velocity field for the discrete bin space is given by:
\begin{equation}
    u^i_t(b^{i}, b \mid \mathcal{P}) = \frac{\dot{\kappa}_t}{\kappa_t} \, \delta_{b}(b^{i}) - p_{0|t}(b^{i} \mid b, \mathcal{P}),
    \label{eq:generating_velocity}
\end{equation}
with \(p_{0|t}(b^{i} \mid b, \mathcal{P}) = \sum_{b_0, b_1 \in \mathcal{B}} \delta_{b_0}(b^{i}) \, p_t(b_0, b_1 \mid b, \mathcal{P})\) denoting the conditional probability of the clean bin \(b_0\) given the noisy observation at time \(t\), and \(\kappa_t\) a time-dependent scaling function determined by the noise schedule.

The model outputs logits over the discrete bin set \(\mathcal{B}\), which are normalized via softmax to yield the final predictive distribution \(p(b \mid \mathcal{P})\). Building on this probabilistic output, we next describe how to sample from the latent space to bridge the discrete predictions with continuous refinement.


\subsection{Anchor–Probs Latent Pose Sampling}
From the predicted distribution $p(b \mid \mathcal{P})$, we select the top-$N$ highest-probability bins:
\begin{equation}
\mathcal{B}_{\text{top-}N} = \left\{ B_{\text{latent}}^{(1)}, B_{\text{latent}}^{(2)}, \dots, B_{\text{latent}}^{(N)} \right\},
\label{eq:tokN}
\end{equation}
where the most probable bin $B_{\text{anchor}} = B_{\text{latent}}^{(1)}$ acts as the \textbf{anchor bin} with confidence $\pi_{\text{anchor}} = p(B_{\text{anchor}} \mid \mathcal{P})$, and the remaining bins $\{B_{\text{latent}}^{(j)}\}_{j=2}^N$ form the \textbf{probs bins} with weights $\pi_j = p(B_{\text{latent}}^{(j)} \mid \mathcal{P})$.

The \textit{anchor-probs} framework balances high-confidence initialization with uncertainty-aware exploration. Unlike multi-hypothesis methods plagued by training ambiguity and high overhead, our strategy ensures stable convergence and efficient exploration via flexible sampling. The top-1 anchor bin serves as the initial pose for subsequent refinement:
\begin{equation}
\hat{\mathbf{P}}_{\text{anchor}} = (\hat{\mathbf{R}}_{\text{anchor}}, \hat{\mathbf{T}}_{\text{anchor}}),
\label{eq:anchor_pose}
\end{equation}
which canonicalizes the input point cloud for refinement. The probs bins and their probabilities define an uncertainty-aware initialization distribution, supporting multi-hypothesis optimization at sub-bin resolutions in the subsequent stage.

To leverage this for continuous refinement, as shown in Fig.~\ref{fig:framework}(b), we employ an \textbf{Adaptive Latent Pose Sampling} strategy. We define the 6D pose residual as $\mathbf{r} = (r^1, \dots, r^6) \in \mathbb{R}^6$, which encapsulates the fine-grained offsets for each pose dimension relative to the discretized bin centers. Specifically, $r^i$ denotes the residual corresponding to the $i$-th dimension, derived from the top-$N$ scoring bins $\{B^{(m)}\}_{m=1}^{N}$ with associated confidences $\{\pi_m\}$. The initial residual distribution is modeled as a Gaussian mixture:
\begin{equation}
\mathbf{r}_0 \sim p_{\mathrm{latent}}(\mathbf{r})  = \prod_{i=1}^{6} p(r^i), p(r^i) = \sum_{m=1}^{N} \pi_m^i \, \mathcal{N}(r^i \mid \mu_{m}^i, \sigma_{m}^i),
\end{equation}
where $\boldsymbol{\mu}_m$ and $\boldsymbol{\Sigma}_m$ capture the center and uncertainty (determined by the bin width) of each bin.
Specifically, sampling between bins is performed based on their probabilities, while Gaussian sampling is adopted within each bin. This approach maintains multi-modality and restricts sampling to promising residual regions, setting the stage for precise pose regression.

\subsection{Continuous Pose Optimization via Flow Matching}

\paragraph{Latent-Guided Residual Flow Training}
To achieve precise pose Optimization in the local latent space, we utilize Continuous Flow Matching (CFM)~\cite{lipman2022flow} to map from the structured initialization distribution to the ground-truth residual $\mathbf{r_1} = (\Delta\mathbf{R}, \Delta\mathbf{T}) \in \mathbb{R}^6$. Using the anchor pose $(\hat{\mathbf{R}}_{\text{anchor}}, \hat{\mathbf{T}}_{\text{anchor}})$ from the latent prediction, we transform the input point cloud to a canonical space:
\begin{equation}
\mathcal{P}_{\text{inv}} = \hat{\mathbf{R}}_{\text{anchor}}^{-1} \left( \mathcal{P} - \hat{\mathbf{T}}_{\text{anchor}} \right),
\label{eq:canonical_transform}
\end{equation}
where residuals are primarily due to discretization, ensuring a localized and stable refinement process. Compared to diffusion models, CFM enables faster inference via deterministic ODE solving and superior sample quality through direct optimal transport. Relative to direct MLP methods, CFM better captures continuous, multi-modal distributions in the residual space, boosting accuracy and robustness without oversimplifying assumptions.

Following CFM's optimal transport framework~\cite{lipman2022flow}, the conditional path from $\mathbf{r}_0 \sim p_{\mathrm{latent}}(\mathbf{r})$ to the target residual $\mathbf{r}_1 \sim q(\mathbf{r} \mid \mathcal{P}_{\text{inv}})$ is:
\begin{equation}
p_t(\mathbf{r} \mid \mathbf{r}_1) = \mathcal{N}\!\left( \mathbf{r} \,\big|\, t\mathbf{r}_1,\; \big[1 - (1 - \sigma_{\min}) t\big]^2 \mathbf{I} \right),
\end{equation}
with transport map $\psi_t(\mathbf{r}_0) = \big[1 - (1 - \sigma_{\min}) t\big] \mathbf{r}_0 + t \mathbf{r}_1$. Training minimizes the flow matching loss:
 
\begin{equation}
\mathcal{L}_{\text{CFM}}(\theta) = \mathbb{E}_{t, \mathbf{r}_0, \mathbf{r}_1} \left[ \left\| v_t\bigl(\psi_t(\mathbf{r}_0); \theta \bigr) - \frac{d}{dt} \psi_t(\mathbf{r}_0) \right\|^2 \right],
\end{equation}
where $t \sim \mathcal{U}[0,1]$, $\mathbf{r}_0 \sim p_{\mathrm{latent}}(\mathbf{r})$, and $\mathbf{r}_1 \sim q(\mathbf{r} \mid \mathcal{P}_{\text{inv}})$ is the ground-truth residual. This aligns the velocity field $v_t$ with optimal trajectories for stable, accurate learning.

\paragraph{Residual Inference and Pose Reconstruction}
At inference, an initial residual $\mathbf{r}_0$ is adaptive sampled from $p_{\mathrm{latent}}(\mathbf{r})$ and evolved using an ODE solver:
\begin{equation}
\frac{d}{dt} \mathbf{r}(t) = v_t\bigl(\mathbf{r}(t); \theta\bigr), \quad \mathbf{r}(0) = \mathbf{r}_0,
\end{equation}
producing the refined residual $\hat{\mathbf{r}} = \mathbf{r}(1) = (\Delta \hat{\mathbf{R}}, \Delta \hat{\mathbf{T}})$. The final pose composes this with the anchor:
\begin{equation}
\hat{\mathbf{R}}_{\mathrm{final}} = \Delta \hat{\mathbf{R}} \otimes \hat{\mathbf{R}}_{\mathrm{anchor}}, \qquad
\hat{\mathbf{T}}_{\mathrm{final}} = \Delta \hat{\mathbf{T}} + \hat{\mathbf{T}}_{\mathrm{anchor}},
\end{equation}
where $\otimes$ denotes $\mathrm{SO}(3)$ rotation composition.

Leveraging canonicalized point clouds and uncertainty-aware initialization, the CFM model efficiently explores sub-bin hypotheses while focusing on high-confidence regions, achieving a balance of multimodality, robustness, and precision for accurate continuous pose refinement.

\paragraph{Articulated objects Extension}
Following the joint-centric strategy~\cite{zhang2025r}, the root part pose $(\hat{\mathbf{R}}_{\text{final}}, \hat{\mathbf{T}}_{\text{final}})$ is localized using the aforementioned two-stage pipeline, the poses of dependent child nodes are resolved by predicting intrinsic joint parameters: the axis $\mathbf{u} \in \mathbb{R}^3$, origin $\mathbf{q} \in \mathbb{R}^3$, and state $s \in \mathbb{R}$. For revolute joints, the relative transformation can be directly derived from the predicted joint parameters (axis, origin, and state) using the formulation and loss in~\cite{zhang2025r}. The pose of each child node is then obtained by applying this transformation along the kinematic chain.

\section{Experiments} \label{sec:Experiments}

\begin{table*}[t!]
\caption{
\textbf{Quantitative comparison on the REAL275 dataset for category-level pose estimation.}
Baseline results are taken from the original papers. $\uparrow$ ($\downarrow$) indicates higher (lower) is better. \textbf{Data} denotes the input format, and \textbf{Prior} indicates the use of category priors. -' indicates that the metrics are not reported in the original paper.
}
\label{table: baselines}
\centering
\resizebox{0.88\textwidth}{!}{
\begin{tabular}{c|c|cc|cccc|c}
\toprule
\multicolumn{2}{c|}{Method}                          & Data  & Prior & $5^{\circ}2$cm$\uparrow$ & $5^{\circ}5$cm$\uparrow$ & $10^{\circ}2$cm$\uparrow$ & $10^{\circ}5$cm$\uparrow$ & Inference Time (s)$\downarrow$ \\ \midrule
              & NOCS~\cite{wang2019normalized}        & RGB-D & $\times$   & -    & 9.5  & 13.8  & 26.7  & -          \\
              & i2c-net~\cite{remus2023i2c}& RGB-D & $\times$  & - & 24.62 & -  &  -  &  0.068 \\
              & SGPA~\cite{chen2021sgpa}              & RGB-D & \checkmark & 35.9 & 39.6 & 61.3  & 70.7  & -          \\
Deterministic & DPDN~\cite{lin2022category}           & RGB-D & \checkmark & 46.0 & 50.7 & 70.4  & 78.4  & -          \\ 
\cmidrule{2-9}
              & GPV-Pose~\cite{di2022gpv}             & D     & $\times$   & 36.2 & 41.1 & -     & 74.2  & 0.050         \\ 
              & Nie et al.~\cite{nie2023category}& D     & \checkmark   & 36.2 & 41.1 & -     & 74.2  & -       \\
              & Tu et al.~\cite{tu2025language}& D     & \checkmark   & 50.3 & 59.5 & 69.7  & 82.5  & -      \\
              & RBP-Pose~\cite{zhang2022rbp}          & D     & \checkmark & 38.2 & 48.1 & 63.1  & 79.2  & 0.040          \\ 
              \midrule \midrule
              & Genpose~\cite{zhang2023generative}       & D     & $\times$   & 52.1 & 60.9 & 72.4  & 84.0  & 0.058        \\
Probabilistic 
            & \textbf{Flow6D}(Ours)        & \textbf{D}     &  \textbf{$\times$}   & \textbf{55.2} & \textbf{64.5} & \textbf{76.3}  &  \textbf{86.1}  & \textbf{0.011}        \\
\bottomrule

\end{tabular}
}

\end{table*}

\begin{figure*}[t!]
    \centering
    \includegraphics[width=\linewidth]{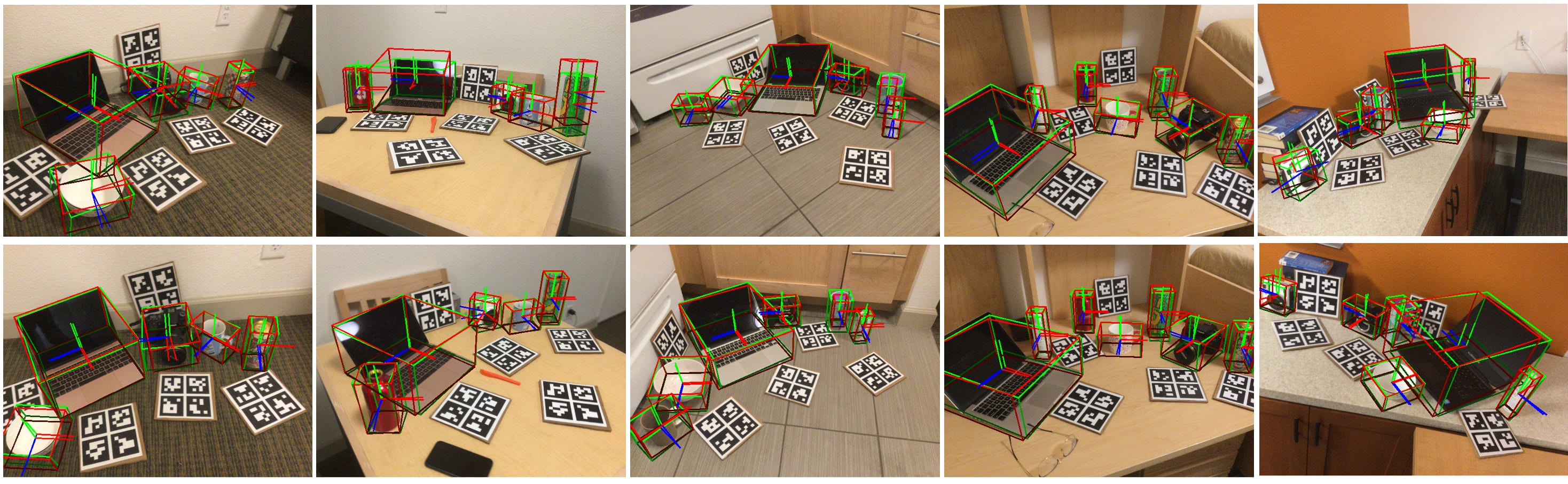}
    \caption{\textbf{Results on the real-world REAL275 Datasets, and red and green 3D boxes represent ground truth and our predictions, respectively.} }
    \vspace{-0.5em}
    \label{fig:Real275}
\end{figure*} 

\subsection{Experimental Settings}

\noindent \textbf{Dataset.}\quad Our method is designed to handle both rigid and articulated objects and is evaluated on a diverse set of synthetic and real-world datasets. Concretely, CAMERA25~\cite{wang2019normalized} and ArtImage~\cite{xue2021omad} are used for evaluation of the synthetic dataset. 
REAL275~\cite{wang2019normalized} and RobotArm~\cite{liu2022toward} are used for evaluation of real-world scenarios.

\noindent \textbf{Metric.} Following prior work~\cite{zhang2025gapt, zhang2025u,zhang2023generative},  we report the rotation error in degrees, the translation error measured in the normalized part coordinate space for each rigid part. To ensure fair comparison across parts of different physical scales, the translation error is further normalized on a per-part basis.

\noindent \textbf{Implementation.} {We conducted simulation and real-robot experiments on articulated objects and rigid bodies. For articulated objects, we sampled 2,048 points, while rigid bodies used 1,024 input points. In real-robot experiments, point clouds were captured with an Intel RealSense D435 RGB-D camera (640×480 at 30 Hz). Following prior protocols, we employed instance masks generated by Mask R-CNN~\cite{he2017mask} during inference. The initial learning rate was set to \(5 \times 10^{-4}\) with a cosine annealing schedule. For both discrete and continuous flow matching, a time-step size of 0.1 was used during inference. For each object category, a category-specific model is trained for pose estimation.} All experiments are implemented in PyTorch and conducted on a single RTX 4090 GPU with a batch size of 128.

\begin{table*}[t!]
    \caption{Comparison with SOTA methods on the ArtImage dataset. We validate our method on categories Laptop, Eyeglasses, Dishwasher, Scissors, and Drawer that contains 2, 3, 2, 2, 4 parts respectively.}
    \label{tab:main}
    \small
    \centering
    \resizebox{0.72\linewidth}{!}{
    \begin{tabular}{c|l|cc|c}
    \toprule
    \multirow{2}{*}{Category} & \multirow{2}{*}{Method} & \multicolumn{2}{c|}{Per-part 6D Pose} & \multirow{2}{*}{Inference Time (s)} \\
    \cline{3-4}
    & & Rotation Error ($^{\circ}$) & Translation Error (m) \\
    \hline
    \multirow{5}{*}{Laptop} & A-NCSH \cite{li2020category} &  5.3, 5.4 & 0.054, 0.043  & 9.0  \\
    & ArtPERL \cite{liu2023categoryRL} & 4.9, 4.7 & 0.053, 0.066  & 0.9 \\
    & U-COPE~\cite{zhang2025u} & 4.8, 4.1 & 0.029, 0.030 & 1.8 \\
    & CAPTRA \cite{weng2021captra} & 5.9, 5.3 & 0.080, 0.063 & 0.10 \\
    & \textbf{Flow6D} (Ours)  & \textbf{2.7, 3.2} & \textbf{0.034, 0.040} & \textbf{0.013} \\
    \hline
    
    \multirow{5}{*}{Eyeglasses} & A-NCSH \cite{li2020category} & 3.7, 22.3, 23.2 & 0.049, 0.313, 0.324  & 11.9 \\
    & ArtPERL \cite{liu2023categoryRL} & 4.1, 6.2, 6.0 & 0.047, 0.095, 0.091  & 1.0 \\
    & U-COPE~\cite{zhang2025u} & 3.9, 5.3, 5.6 & 0.043, 0.088,  0.088 & 2.1 \\
    & CAPTRA \cite{weng2021captra} & 4.5, 12.6, 13.1 & 0.054, 0.097, 0.084 & 0.14 \\
    & \textbf{Flow6D} (Ours)  & \textbf{3.1, 4.8, 4.9} & \textbf{0.036, 0.080, 0.083} & \textbf{0.015} \\
    \hline
    
    \multirow{5}{*}{Dishwasher} & A-NCSH \cite{li2020category} & 4.0, 4.8 & 0.059, 0.123 & 5.5 \\
    & ArtPERL \cite{liu2023categoryRL} & 3.9, 4.3 & 0.055, 0.079  & 0.9 \\
    & U-COPE~\cite{zhang2025u} & 3.8, 4.5 & 0.062, 0.066  & 1.4 \\
    & CAPTRA \cite{weng2021captra} & 4.6, 5.4 & 0.055, 0.089 & 0.11 \\
    & \textbf{Flow6D} (Ours)  & \textbf{3.0, 3.5} & \textbf{0.040, 0.044} & \textbf{0.014} \\
    \hline
    
    \multirow{5}{*}{Scissors} & A-NCSH \cite{li2020category} & 2.0, 2.9 & 0.035, 0.025  & 6.5 \\
    & ArtPERL \cite{liu2023categoryRL} & 2.2, 2.6 & 0.031, 0.042  & 0.8 \\
    & U-COPE~\cite{zhang2025u} & 2.4, 2.5 & 0.033, 0.023 & 1.9 \\
    & CAPTRA \cite{weng2021captra} & 4.1, 4.7 & 0.032, 0.039 & 0.12 \\
    & \textbf{Flow6D} (Ours)  & \textbf{1.8, 2.2} & \textbf{0.019, 0.022} & \textbf{0.013} \\
    \hline
    
    \multirow{5}{*}{Drawer} & A-NCSH \cite{li2020category} & 2.8, 3.5, 3.9, 2.9 & 0.045, 0.155, 0.157, \textbf{0.075}  & 16.5\\
    & ArtPERL \cite{liu2023categoryRL} & 3.5, 3.5, 3.5, 3.5 & 0.061, 0.112, 0.121, 0.104 & 1.1 \\
    & U-COPE~\cite{zhang2025u} & 2.7, 3.2, 3.4, 2.9 & 0.042, 0.101, 0.122, 0.094 & 1.7 \\ 
    & CAPTRA \cite{weng2021captra} & 4.8, 6.5, 6.3, 6.0 & 0.112, 0.185, 0.177, 0.156 & 0.25 \\
    & \textbf{Flow6D} (Ours) & \textbf{2.4, 2.4, 2.4, 2.4} & \textbf{0.044, 0.080, 0.091,} 0.077 & \textbf{0.019} \\ 
    \bottomrule
    \end{tabular}}

\end{table*}

\subsection{Experimental Results}
In this work, to ensure fairness and comprehensiveness in experimental comparisons, we select representative benchmark datasets separately for rigid objects and articulated objects. Specifically, the evaluation setup for rigid objects follows the protocol of Genpose~\cite{zhang2023generative}, while the evaluation protocol and data splits for articulated objects adhere to the standards adopted in U-COPE~\cite{zhang2025u}. This design ensures that different methods are compared in a consistent and appropriate manner within their respective task settings.

\noindent \textbf{Quantitative Results.}
Tab~\ref{table: baselines} and Tab~\ref{tab:main} present the quantitative performance comparison of the proposed method with other baseline methods, including rotation error and normalized translation error. For rigid objects, Flow6D achieves optimal performance using only depth information (without prior), which can be attributed to the proposed structured efficient searching.
For articulated objects, Flow6D achieves significant performance advantages. Particularly in the rotation errors, it reduces by 2.6$^\circ$ and 2.2$^\circ$ respectively in \textit{Laptop}, compared to A-NCSH, fully verifying the effectiveness of the latent space localization to continuous regression hierarchical strategy proposed in this paper. Discrete latent space localization greatly reduces the search space, and continuous pose optimization accurately offsets discretization errors, which together enhance the accuracy of pose estimation. Meanwhile, the method outperforms other baselines in  AP metrics, demonstrating unified adaptability to both rigid and articulated objects. Notably, 
the inference time in Table~\ref{table: baselines} is averaged across all categories, since variations between rigid objects are minimal. With an inference speed of less than 0.020$s$, our method surpasses all comparison methods, successfully balancing accuracy and efficiency and providing a reliable solution for real-time 6D pose estimation scenarios.

\begin{figure}[t!]
    \centering
    \includegraphics[width=0.98\linewidth]{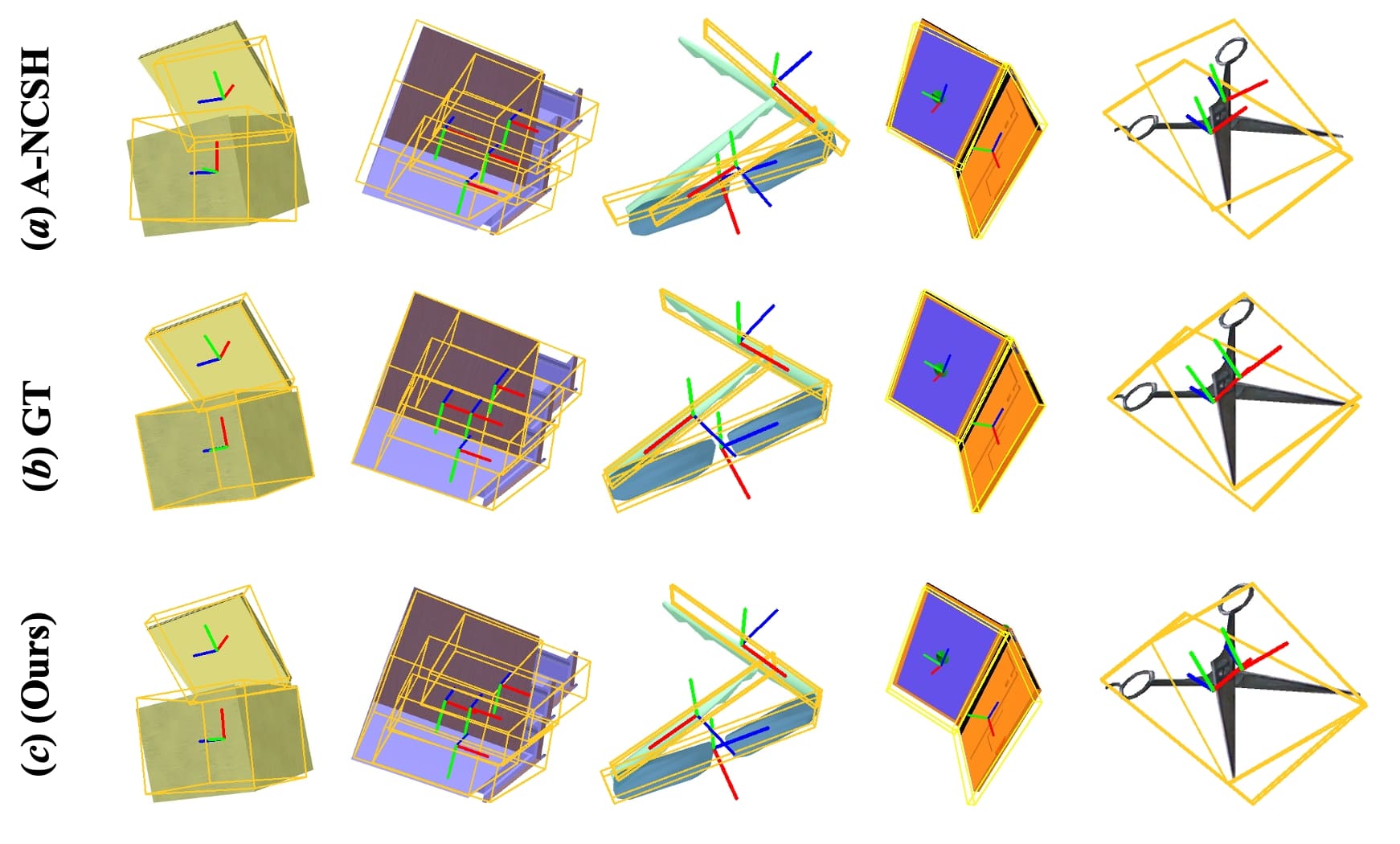}
    \caption{\textbf{Results on the ArtImage Dataset.} }
    \vspace{-0.5em}
    \label{fig:artimage}
\end{figure}

\noindent \textbf{Qualitative Results.} 
Fig.~\ref{fig:Real275} and Fig.~\ref{fig:artimage} present qualitative comparison results of the proposed method on synthetic and real-world datasets, covering complex scenarios (e.g., occlusion, noise interference) for both rigid and articulated objects. It can be observed that Flow6D accurately estimates the 6D pose of targets, with predicted 3D bounding boxes highly aligned with ground truth and significant calibration effects on rotation and translation. Compared to baselines, our method achieves more accurate pose estimation at object joint connections and locally occluded regions, effectively avoiding pose deviation issues that baseline methods are prone to. This benefit stems from the latent space-to-pose regression hierarchical strategy: discrete latent bin prediction locks the approximate pose range, while continuous fine refinement corrects subtle deviations, enabling the model to capture core pose features even in complex scenarios. Qualitative results further validate the method’s effectiveness, and its output pose results can provide reliable support for practical applications such as robotic grasping.

\subsection{Ablation Study}
We conduct experiments on the ArtImage dataset (base part of the \textit{Laptop} category) to evaluate the impact of different design choices in our two-stage framework.

\begin{table}[!t]
\caption{
\textbf{Ablation Study of Discrete Bin in Stage I and Latent Top-$N$ Bin Selection in Stage II.}
GT-in-Top-$N$ indicates whether the ground-truth rotation or translation bin is included among the Top-$N$ predictions.
``--'' denotes using only the most probable bin center without Stage II
}
\label{tab:ablation_supp}
\centering
\resizebox{\linewidth}{!}{
\begin{tabular}{c|c|c|c|c|c}
\toprule
\textbf{Stage I} & \textbf{Stage II} & \multicolumn{2}{c|}{\textbf{GT-in-TopN Rate (\%)}} & \multicolumn{2}{c}{\textbf{Final 6D Pose Error}} \\
\textbf{Bin Size} & \textbf{Latent Bin size} & \textbf{Rotation.} & \textbf{Translation.} & \textbf{Rot. Err. ($^\circ$)} & \textbf{Trans. Err. (m)} \\
\midrule
\multirow{4}{*}{24} 
& --    & -- & -- & 8.9 & 0.078 \\
& 1     & 92.1 & 32.8 & 7.5 & 0.069 \\
& 3     & 99.2 & 78.2 & 6.9 & 0.060 \\
& 5     & 99.5 & 95.0 & 6.4 & 0.055 \\
\midrule
\multirow{6}{*}{36} 
& --    & -- & -- & 6.7 & 0.052 \\
& 1     & 85.0 & 23.1 & 5.2 & 0.048 \\
& 3     & 98.9 & 70.3 & 4.4 & 0.042 \\
& 5     & 99.4 & 91.4 & 3.9 & 0.038 \\
& 10    & 99.7 & 99.3 & 4.2 & 0.043 \\
\midrule
\multirow{6}{*}{60} 
& --    & -- & -- & 4.7 & 0.046 \\
& 1     & 90.3 & 28.6 & 3.5 & 0.041 \\
& 3     & 98.9 & 76.6 & 3.1 & 0.037 \\
& \textbf{\underline{5}}     & 99.7 & 92.3 & \textbf{\underline{2.7}} & \textbf{\underline{0.034}} \\
& 10    & 99.9 & 99.1 & 3.3 & 0.038 \\
\midrule
\multirow{4}{*}{100} 
& --    & -- & -- & 4.6 & 0.050 \\
& 1     & 64.2 & 10.1 & 4.3 & 0.047 \\
& 3     & 77.7 & 52.5 & 4.1 & 0.044 \\
& 5     & 89.9 & 85.1 & 3.9 & 0.041 \\
\bottomrule
\end{tabular}
}
\vspace{-0.5em}
\end{table}

\textbf{Discrete Bin Size.} The discrete flow matching model for latent space localization relies on discretizing pose parameters into bins. The bin size directly influences both latent localization accuracy and the efficiency of subsequent refinement. To analyze this effect, we conduct ablation studies on the bin size, with results reported in Tab.~\ref{tab:ablation_supp}. The setting labeled as the latent bin size “--''  corresponds to predicting pose only stageI.
Experimental results indicate that a bin size of 60 achieves the best overall performance, yielding the lowest rotation and translation errors. Smaller bin sizes (e.g., 24 or 36) result in coarse discretization and large initial errors, hindering effective refinement and convergence. Increasing the bin size beyond 60, for example to 100, leads to only marginal accuracy gains while incurring higher computational cost and longer inference time. These results confirm that a bin size of 60 provides an effective balance, enabling reliable latent space localization while preserving sufficient space for pose refinement.

\textbf{Latent Bin Size.}
Latent Bin Size defines the latent search space for Stage II pose regression. As shown in Table~\ref{tab:ablation_supp}, disabling Stage II (``--'') or using a single bin ($N=1$) makes final accuracy strongly dependent on Stage I predictions. Although rotation recall remains high (90.3\% at Bin Size 60), translation recall is low (28.6\%) due to weak discrimination among neighboring translation bins. Increasing the latent bin size expands the refinement space and improves pose coverage. A moderate setting ($N=5$) provides the best trade-off, achieving high GT-in-Top-$N$ rates and the lowest pose errors, while larger $N$ introduces excessive residual space that degrades convergence and accuracy.

 Overall, \textbf{Latent Bin Size 5, with Bin Size 60}, provides the optimal balance between localization coverage and optimization stability. Despite high Top-5 coverage (99.7\% rot., 92.3\% trans.), Stage II retains an \textbf{error-correction capability} that consistently regresses towards the ground truth, even when starting from the boundaries of the latent space. Crucially, our method models six pose dimension independently, preventing error propagation and preserving partial geometric consistency even if a single axis misses the true mode.

\subsection{Inference Speed Analysis}
On an NVIDIA RTX 4090 (batch size 1), our method achieves real-time performance on ArtImage and CAMERA25. Table~\ref{tab:inference_times} summarizes inference times for representative scissors and bottle categories. The inference pipeline achieves a total latency of \(0.012 \pm 0.002\) s per frame, comprising  point cloud feature extraction,  discrete bin prediction, and  continuous pose regression. For articulated objects, predicting the child-part pose incurs an additional ~0.002 s, yielding a stable processing rate of approximately 70 FPS. 
\begin{table}[h!]
    \caption{\textbf{Inference speed analysis} on two dataset categories.}
    \label{tab:inference_times}
    \centering
    \begin{tabular}{l r r}
        \toprule
        \textbf{Module} & \textbf{ArtImage(\textit{scissors})} & \textbf{CAMERA25(\textit{bottle})} \\
        \midrule
        Point cloud encoder               & 0.0078 s & 0.0073 s \\
        Latent bin prediction             & 0.0021 s & 0.0023 s \\
        pose regression                   & 0.0015 s & 0.0017 s \\
        Child part prediction             & 0.0018 s & -- \\
        \midrule
        Total inference time              & 0.0133 s & 0.0113 s \\
        \bottomrule
    \end{tabular}

\end{table}

\begin{figure}[t!]
    \centering

    \includegraphics[width=\linewidth]{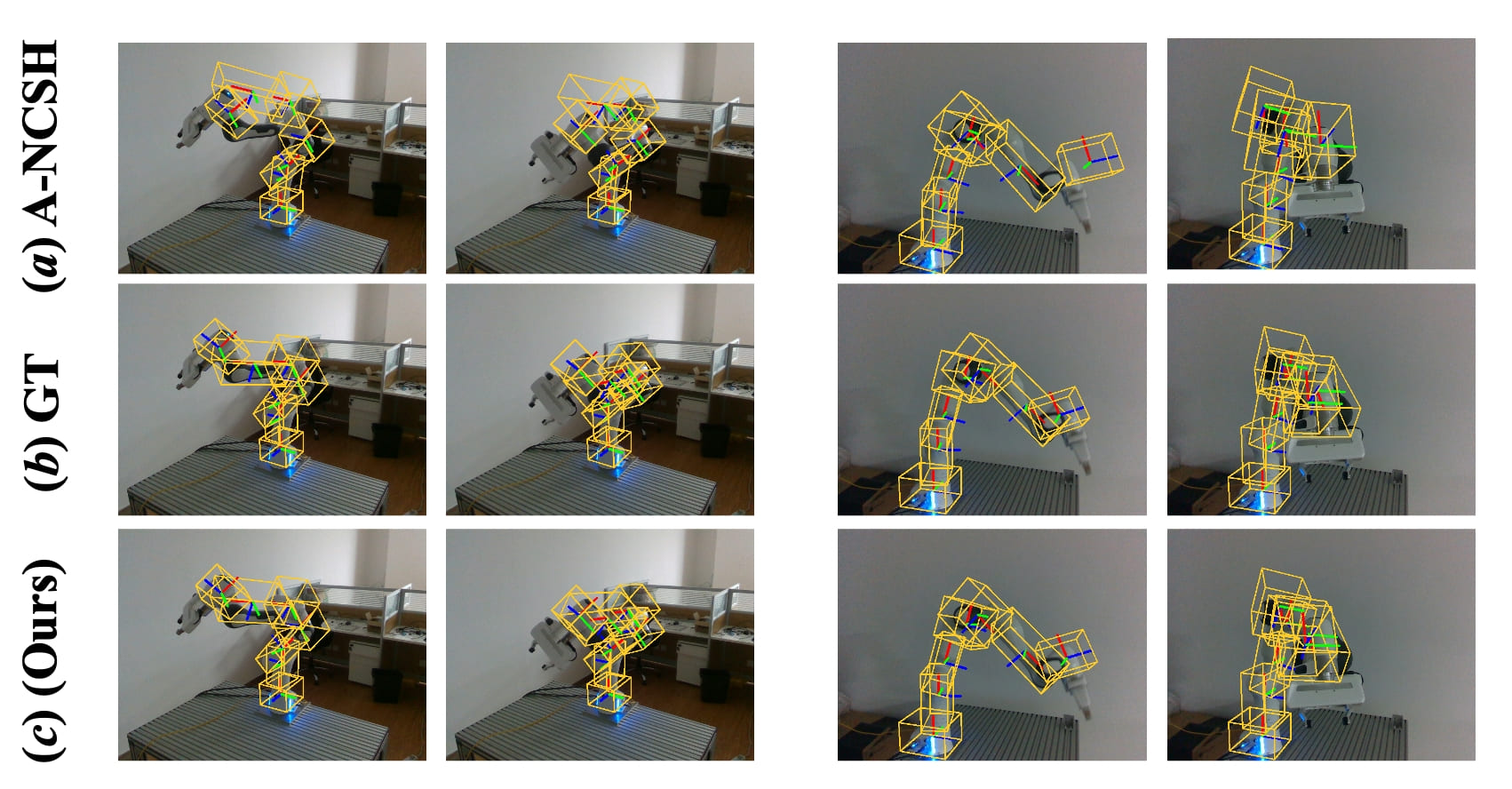}
    \caption{\textbf{Results on the real-world RoboArm Dataset.} }
    \label{fig:art_realworld}
\end{figure} 

\subsection{Generalization Capacity on Real World}

To verify the proposed method’s performance in real-world scenarios, we conduct tests on the RobotArm~\cite{liu2022toward} and REAL275~\cite{wang2019normalized} dataset. REAL275 contains rich real-environment interferences (e.g., illumination changes, object occlusion, cluttered backgrounds), which effectively evaluates the model’s practical adaptability. Experimental results Fig.~\ref{fig:Real275} show that Flow6D still maintains excellent performance in real scenes: the predicted 6D pose is highly consistent with the target’s ground truth, the 3D bounding box accurately fits the object contour, and both rotation and translation errors are controlled within practical ranges. 

For fina dataset, We  evaluate Flow6D using the 7-part RobotArm dataset in real-world scenarios ,the results Fig.~\ref{fig:art_realworld} demonstrate the robustness and accuracy of our approach in handling complex, multi-part articulations in realworld settings.
These results fully demonstrate the method’s strong robustness to real-world interferences, providing solid support for practical application deployment.

\section{CONCLUSIONS}

In this paper, we employ flow matching for category-level pose estimation. Through a novel two-stage framework that bridges discrete and continuous representations, our method achieves high precision and exceptional speed, attaining state-of-the-art (SOTA) performance on multiple datasets. However, our method has not been validated under mutual occlusions in multi-object scenes. Furthermore, unifying this two-stage approach into an end-to-end model is a promising avenue for future research.








\printbibliography

\end{document}